\documentclass{article}

\usepackage{CJKutf8}
\usepackage{arxiv}
\usepackage{subfigure}
\usepackage{color}
\usepackage[utf8]{inputenc} 
\usepackage[T1]{fontenc}    
\usepackage{hyperref}       
\usepackage{url}            
\usepackage{booktabs}       
\usepackage{amsfonts}       
\usepackage{nicefrac}       
\usepackage{microtype}      
\usepackage{lipsum}
\usepackage{graphicx}
\usepackage{mwe} 
\usepackage{wrapfig}
\graphicspath{ {./images/} }

\title{JPS-daprinfo: A Dataset for Japanese Dialog Act Analysis and People-related Information Detection}

\author{\large
 Changzeng Fu$^{1,2}$ \\ 
  $^{1}$Graduate School of Engineering Science, Osaka University, Japan\\
  $^{2}$Advanced Telecommunications Research Institute International, Japan\\
  \texttt{changzeng.fu@irl.sys.es.osaka-u.ac.jp} \\
 }

\begin{document}
\maketitle


\large

\section{Dataset Description}

Robots’ or smart agents' memory construction and sharing have been proved to be important in human-robot interaction (HRI) and human-agent interaction (HCI) by many studies \cite{fu2020sharing, mitsuno2020robot, mahzoon2019effect}. These studies suggest that regarding people-related information as a type of memory and sharing it with the user can be a good behavior for maintaining interaction. However, most of the relevant studies prepared the robot's memories manually, instead of obtaining them from robots’ actual activities. This lack of memory constructing ability makes the strategy of sharing memories less practical.

To build the people-related information detector, we firstly prepared a spoken Japanese dataset with a two-way Japanese conversation (I-JAS \cite{KumikoSAKODA2018international}), which contains 50 interview dialogues of two-way Japanese conversation that discuss the participants’ past present and future. Each dialogue is 30 minutes long.
From this dataset, we selected the interview dialogues of native Japanese speakers as the samples. 
Given the dataset, we annotated sentences \footnote[1]{The dataset for training can be found at \url{https://github.com/CZFuChason/JPS-daprinfo}} (utterance-level) with 13 labels by referring to Stolke et al.'s \cite{stolcke2000dialogue} method with some modifications. The labeling work was conducted by native Japanese speakers who have experiences on the data annotation.
The total amount of the annotated samples is 20130. 
Table \ref{tabel:prsampels} shows the labels, definitions, corresponding samples as well as the amount of each category.

\begin{table}[t]
\centering
\caption{Samples of the dataset}
\begin{tabular}{p{3.5cm}p{4.5cm}p{6.5cm}p{1cm}}
\toprule

Label & Definition & Example & Amount\\
\noalign{\smallskip}\hline\noalign{\smallskip}
subjective information & subjective opinions toward something & \begin{CJK}{UTF8}{min}このカップラーメンとてもおいしいです。\end{CJK}// This cup noodle is delicious. & 5707\\

objective information & a fact-based statement toward something & \begin{CJK}{UTF8}{min}今日はお隣のスーパーでお弁当を、買って来て食べました。
\end{CJK}// Today, I had my lunch with a lunch box bought from a nearby store.  & 2752\\

plan & a temporal statement of intended actions & \begin{CJK}{UTF8}{min}なんか十月から、一か月半ロンドンに行く予定があって。
\end{CJK}// Well, I am planning to go to London for a month and a half in October.  & 54\\

question & an utterance which typically functions as a request for information & \begin{CJK}{UTF8}{min}将来的に、住むとしたら、田舎と都会と、どちらがいいと思いますか。
\end{CJK}// Where do you prefer to stay in the future, the country or the city?  & 3457\\

apology & an expression of regret & \begin{CJK}{UTF8}{min}すみません怖い話を思い出させてしまいました。\end{CJK}// I am sorry for reminding you of a scary story.  & 96\\

thanking & an expression of gratitude & \begin{CJK}{UTF8}{min}あー、ありがとうございます、いいい話。\end{CJK}// Oh, thank you, a good talk. & 343\\

topic changing/closing & an expressions to change or end a topic & \begin{CJK}{UTF8}{min}じゃあ、あのー最後の質問なんですけど。\end{CJK}// Okay, well, I have one last question.  & 330\\

agreement & an expressions for feeling or thinking the same way about something & \begin{CJK}{UTF8}{min}はい、そうですねー。\end{CJK}// Yes, it is.  & 1917\\

disagreement & an expressions for feeling or thinking in a different way about something & \begin{CJK}{UTF8}{min}それは、ない、ですね。\end{CJK}// Um, it is not.  & 94\\

request & an expressions for asking something & \begin{CJK}{UTF8}{min}お弁当をください。\end{CJK}// Please give me a lunch box.  & 239\\

proposal & a suggestion for doing something & \begin{CJK}{UTF8}{min}コンビニに行きましょか。\end{CJK}// Let's go to the convenience store.  & 58\\

summarize/reformulate &  a brief statement of the main points & \begin{CJK}{UTF8}{min}あーなるほどね、あじゃあ鳥取といえば水木しげるっていう感じなんですね。\end{CJK}// Oh, I see, so Tottori is known for Shigeru Mizuki.  & 1999\\

other & backchannel, filler, echo, etc. & \begin{CJK}{UTF8}{min}あ、へー / うん / んー / なるほど / あー /そうですかー。\end{CJK}// oh, well / hmm / ah / yeah-.  & 3985\\

\bottomrule
\label{tabel:prsampels}
\end{tabular}
\end{table}

\section{Text Classification}
We simply run a text classification based on SeMemNN \cite{fu2020sememnn} and BERT\cite{devlin2018bert}. The training-set and testing-set were split as predefined.
Here are the results:
\begin{itemize}
\large
\item SeMemNN: 76.32\%(UA), 75.41\%(WA);
\item BERT: 77.16\%(UA), 79.33\%(WA);
\end{itemize}
The results show that BERT outperformed SeMemNN. However, the BERT needs a pretraining procedure with a large amount of dataset (the Japanese wiki was used) and the amount of trainable parameters is significantly more than SeMemNN. when the pre-training data or the computing resource is insufficient, SeMemNN may be a better choice.

\section{Potential Usage of the dataset}
In addition to the people-related information detection, there are many potential usages of this annotated dataset.
It is known that the ability to mode and automatically detect discourse structure is an important step toward understanding spontaneous dialogue. This dataset can also be used as a source of Japanese conversation data for such studies. 
Moreover, the labels could also benefit the human-machine conversation if the machine can detect the users' dialogue action well, and use it as auxiliary information to give a response back.
Furthermore, this data can also be used to explore the data unbalance problem on text classification tasks.


\bibliographystyle{IEEEtran}
\bibliography{references}

\end{document}